\documentclass[letterpaper, 10 pt, conference]{ieeeconf}  

\IEEEoverridecommandlockouts                              

\usepackage[printwatermark]{xwatermark}
\usepackage{xcolor}
\usepackage{multirow}
\usepackage{amsmath}
\usepackage{amsthm}
\usepackage{amssymb}
\usepackage{graphicx} 
\usepackage{subfig}
\usepackage{booktabs}
\usepackage[ruled]{algorithm2e}
\usepackage{flushend}
\usepackage[font=small,skip=0pt]{caption}
\usepackage{cite}
\usepackage{caption}
\newtheorem{theorem}{Theorem}

\theoremstyle{definition}

\newcommand{\mb}[1]{\mathbf{ #1 }}
\newcommand{\bs}[1]{\boldsymbol{ #1 }}

\usepackage{array}
\newcolumntype{C}[1]{>{\centering\arraybackslash}p{#1}}
\begin{document}

\title{\LARGE \bf
Quadrotor Trajectory Tracking with Learned Dynamics: Joint Koopman-based Learning of System Models and Function Dictionaries
}

\author{Carl Folkestad$^1$, Skylar X. Wei$^1$, and Joel W. Burdick$^1$\thanks{$^1$All authors are with the Division of Engineering and Applied Sciences, California Institute of Technology, Pasadena, CA, USA $\qquad \qquad$
\texttt{\{cfolkest, swei, jburdick\}@caltech.edu.}}}

\maketitle
\thispagestyle{empty}
\pagestyle{empty}

\begin{abstract}
Nonlinear dynamical effects are crucial to the operation of many agile robotic systems.  Koopman-based model learning methods can capture these nonlinear dynamical system effects in higher dimensional {\em lifted bilinear} models that are amenable to optimal control. However, standard methods that lift the system state using a fixed function dictionary before model learning result in high dimensional models that are intractable for real time control. This paper presents a novel method that jointly learns a function dictionary and lifted bilinear model purely from data by incorporating the Koopman model in a neural network architecture. Nonlinear MPC design utilizing the learned model can be performed readily. We experimentally realized this method on a multirotor drone for agile trajectory tracking at low altitudes where the aerodynamic ground effect influences the system's behavior. Experimental results demonstrate that the learning-based controller achieves similar performance as a nonlinear MPC based on a nominal dynamics model in medium altitude. However, our learning-based system can reliably track trajectories in near-ground flight regimes while the nominal controller crashes due to unmodeled dynamical effects that are captured by our method. 
\end{abstract}

\section{Introduction}
\label{sec:introduction}
The process of designing high performance controllers for agile robotic systems that satisfy state and actuation constraints is challenging for systems with important nonlinear dynamical effects. Model predictive control (MPC) can capture appropriate performance objectives and constraints, and it can be used with \textit{nonlinear} dynamics models if carefully implemented \cite{Kouzoupis2018RecentControl, Gros2020FromIteration, Grandia2020NonlinearFunctions}. However, obtaining a sufficiently accurate dynamical model is {\em crucial} to achieve good performance with nonlinear MPC (NMPC). Many learning approaches have been proposed to capture a robots's complex mechanics and environmental interactions, reducing the need for time consuming system identification (cf. \cite{Rasmussen2018GaussianLearning, Shi2019NeuralDynamics, Taylor2019EpisodicSystems, Recht2019AControl, Kaiser2018SparseLimit}). However, NMPC design based on these models is typically impossible because of the model discretization and the intractable forward simulation needed to evaluate the nonlinear program. Building on our previous work \cite{op_icra_21}, we take a Koopman-centric approach to jointly learn a function dictionary and a \textit{lifted bilinear model} of the dynamics that can readily be used with real-time NMPC to obtain nearly optimal controllers that respect state and actuation constraints. 

We focus on learning control-affine dynamics of the form
    \begin{equation}\label{eq:ca-dynamics}
        \dot{\mb x} = \mb f(\mb x) + \mb g(\mb x)\mb u
    \end{equation} 
where $\mb x$ is the system state and $\mb u$ is a vector of control inputs.  This model characterizes  a wide class of aerial and ground robots. Rather than describing a system's behavior by its state space flows, Koopman methodologies study the evolution of {\em observables}, which are functions over the state-space. In this space, a dynamical system can be represented by a {\em linear} (but possibly infinite dimensional) operator \cite{Lan2013LinearizationSpectrum, Mauroy2016LinearOperator}, which is approximated in practice by a finite dimensional model.  The extended dynamic mode decomposition (EDMD) \cite{Schmid2010DynamicData, Williams2015ADecomposition, Korda2018LinearControl, Brunton2016DiscoveringSystems, Kaiser2021Data-drivenControl, Proctor2018GeneralizingControl} is a common identification technique.

Most current EDMD-methods model the unknown control system dynamics by a lifted \textit{linear} model, which implicitly restricts the control vector fields, $\mb g(\mb x)$, to be state-invariant. This is a significant limitation, as many robotic systems, e.g. systems where input forces enter the dynamics through rotation matrices, are best described by nonlinear control-affine dynamics. Instead, we learn a model underpinned by the \textit{Koopman canonical transform} (KCT) \cite{Surana2016KoopmanSystems}, which allows a large class of nonlinear control-affine dynamic models to be described by a \textit{bilinear} (but possibly infinite dimensional) dynamical system. Our previous work showed how to formulate an EDMD-type method to identify an approximate lifted bilinear model from data and how to utilize the bilinear model for NMPC \cite{op_icra_21}. However, for realistic systems, such as the quadrotor drone considered in the experimental section, the dimension of the lifted model becomes too high when using a fixed function dictionary, resulting in the NMPC being intractable to implement in real-time.

As a result, we incorporate the bilinear Koopman structure in a neural network model, allowing the function dictionary and bilinear model to be learned jointly from data. This enables similar or improved prediction performance with much fewer observable functions, thus allowing real-time use. Additionally, the resulting model maintains the bilinear structure, making it partially interpretable and possible to simulate its behavior forward in time by only evaluating the neural network at the initial state and propagating the lifted state forward using the bilinear system matrices.  Choosing a good function dictionary is a central challenge in Koopman-based methods, and using neural networks to jointly learn Koopman models and function dictionaries has been attempted previously. Multiple works have shown that using an encoder-decoder type architecture to parametrize the function dictionary allows many of the benefits of a Koopman-based model to be maintained while obtaining more compact model and/or improving prediction performance compared to fixed dictionaries \cite{Li2017ExtendedOperator, Kaiser2021Data-drivenControl, Wang2021DeepRacing}. However, no existing method utilizes the proper bilinear models to accurately capture control-affine systems, nor has a view towards achieving high performance control for robotic systems. As such, our contributions are:
\begin{enumerate}
    \item We develop a flexible learning method underpinned by the Koopman canonical transformation to a {\em bilinear} form that can readily be used for NMPC design.
    \item We show in experiments with a quadrotor that a controller based strictly on the data-driven model can outperform an NMPC based on a known, identified model, while needing limited training data.
\end{enumerate}

    
This manuscript is organized as follows. Preliminaries on the KCT are presented in Section \ref{sec:prelims}, and the problem statement and assumptions are discussed in Section \ref{sec:modeling}. Then, the method to jointly learn the function dictionary and bilinear model is presented in Section \ref{sec:learning} before control design is discussed in Section \ref{sec:nmpc}. Finally, the method is demonstrated experimentally on a quadrotor drone in Section \ref{sec:experiments} followed by concluding remarks in Section \ref{sec:conclusion}.

\section{Koopman theory preliminaries}
\label{sec:prelims}

\subsection{Koopman spectral theory}
Before considering the effects of control inputs, we introduce the Koopman operator, which is defined for autonomous continuous-time dynamical systems, $\mb{\dot{x}} = \mb f_\text{aut}(\mb x)$, with state $\mb x \in \mathcal{X} \subset \mathbb{R}^d$, and $\mb f_\text{aut}$ is assumed to be Lipschitz continuous on $\mathcal{X}$. The flow is denoted by $S_t(\mb x)$, defined as $\frac{d}{dt}S_t( \mb x) = \mb f_\text{aut}(S_t(\mb x))$ for all $\mb x \in \mathcal{X}, t\geq 0$. The \textit{Koopman operator semi-group} $(U_t)_{t\geq0}$, from now on simply denoted as the \textit{Koopman operator}, is defined as:
\begin{equation}
    U_t \bs \varphi = \bs \varphi \circ S_t
\end{equation}
for all $\bs \varphi \in \mathcal{C}(\mathcal{X})$, where $\mathcal{C}(\mathcal{X})$ is the space of continuous observables $\bs \varphi: \mathcal{X} \rightarrow \mathbb{C}$, and $\circ$ denotes function composition. Each element of $U_t: \mathcal{C}(\mathcal{X}) \rightarrow \mathcal{C}(\mathcal{X})$ is a \textit{linear} operator. 

An \textit{eigenfunction} of the Koopman operator associated to an eigenvalue $\lambda \in \mathbb{C}$ is any function $\phi \in \mathcal{C}(\mathcal{X})$ that defines a coordinate evolving linearly along the flow of the system:
\begin{equation}
    (U_t \phi)(\mb x) = \phi(S_t(\mb x)) = e^{\lambda t} \phi (\mb x).
\end{equation}


\subsection{The Koopman canonical transform}
We now return to control-affine dynamics (\ref{eq:ca-dynamics}).
and recall when and how they can be transformed to a \textit{bilinear} form through the \textit{Koopman canonical transform} \cite{Surana2016KoopmanSystems}. Let $\big(\lambda_i, \phi_i(\mb x) \big), i = 1,\dots,n$ be eigenvalue-eigenfunction pairs of the Koopman operator associated with the autonomous dynamics of (\ref{eq:ca-dynamics}). The KCT relies on the assumption that the state vector can be described by a finite number of eigenfunctions, i.e. that $ \mb x = \sum_{i=1}^n \phi_i(\mb x) \mb v_i^{\mb x}$ for all $ \mb x \in \mathcal{X}$, and where $\mb v_i^{x} \in \mathbb{C}^d$. This is likely to hold if $n$ is large. If not, they may be well approximated by $n$ eigenfunctions.

When $\phi_i: \mathcal{X} \rightarrow \mathbb{R}$, the KCT is defined as
\begin{align}\label{eq:kct}
        \mb x = C^{\mb x} \mb z, \quad \mb{ \dot{z}} = F \mb z + \sum_{i=1}^mL_{g_i}T(\mb x)u_i
\end{align}
\noindent where $\mb z = T(\mb x) = [\phi_1(\mb x) \dots \phi_n(\mb x)]^T$, $C^{\mb x} = [\mb v_1^{\mb x}\dots \mb v_n^{\mb x}]$, and $F \in \mathbb{R}^{n \times n}$ is a diagonal matrix with entries $F_{i,i} = \lambda_i$. 

Under certain conditions, system (\ref{eq:kct}) is bilinearizable in a countable, possibly infinite basis. We restate the conditions for the existence of such bilinear form in the following theorem that underpins our learning method. 

\begin{theorem} \label{th:kbf}
\cite{Goswami2018GlobalApproach} Suppose there exist Koopman eigenfunctions $\phi_j, j=1,\dots,n$, where $n \in \mathbb{N}, n < \infty$ of the autonomous dynamics of (\ref{eq:ca-dynamics}) whose span, $span(\phi_1,\dots,\phi_n)$, forms an invariant subspace of $L_{g_i}, i = 1,\dots,m$. Then, the system (\ref{eq:ca-dynamics}), and in turn system (\ref{eq:kct}), are bilinearizable with an n-dimensional state space.
\end{theorem}

Hereafter, the finite basis of functions $\{\phi\}$ used in the KCT is termed the {\em function dictionary}.

Although the conditions of Theorem \ref{th:kbf} may be hard to satisfy in a given problem, an approximation of the true system (\ref{eq:ca-dynamics}) can be obtained with sufficiently small approximation error by including adequately many eigenfunctions in the basis. As a result, $L_{g_i} = G_i$ and the system can be expressed as the \textit{Koopman bilinear form} (KBF) (see \cite{Goswami2018GlobalApproach} for details): 
\begin{equation} \label{eq:zdot}
    \mb{\dot{z}}=F \mb z+\sum_{i=1}^m G_i \mb z u_i, \quad \mb z \in \mathbb{R}^n, n < \infty .
\end{equation}

\section{System modeling and data collection}
\label{sec:modeling}


\subsection{Modeling assumptions}

We model the quadrotor dynamics using states global position $\mb p \in \mathbb{R}^3$, velocity $\mb v \in \mathbb{R}^3$, attitude rotation matrix $R \in \text{SO(3)}$, and body angular velocity $\bs \omega \in \mathbb{R}^3$, and consider the following dynamics: 
\begin{equation} \label{eq:dynamics}
\begin{aligned}[c]
&\mb{\dot p} = \mb v,\\
&\dot R = RS(\bs \omega),
\end{aligned}
\quad
\begin{aligned}[c]
& m \mb{\dot v} = m\mb g + R\mb f + \mb f_{\mb v}(\!\begin{smallmatrix}\mb p, \mb v, R, \bs \omega, \mb f, \bs \tau\end{smallmatrix}\!),\\
& J \bs {\dot \omega} = J \bs \omega \times \bs \omega + \bs \tau + \bs \tau_{\bs \omega}(\!\begin{smallmatrix}\mb p, \mb v, R, \bs \omega, \mb f, \bs \tau\end{smallmatrix}\!),
\end{aligned}
\end{equation}
where $m$ and $J$ are the vehicle mass and inertia matrices, respectively, $S(\cdot)$ is a $3\times 3$ skew symmetric mapping, and $\mb g = [0, 0, -g]^T$ is the gravitational force vector. The state
and control-dependent functions $f_{\mb v}(\cdot)$ and $\tau_{\bs \omega}(\cdot)$ capture unmodelled dynamic terms, such as the near-ground effect.

As is common, we abstract the system's control inputs to a total thrust in the body z-direction, $\mb f = [0, 0, T]^T$, and body torques $\bs \tau = [\tau_x, \tau_y, \tau_z]^T$, $\mb u = [T, \tau_x, \tau_y, \tau_z]^T$. The mapping of rotor speeds to the abstract controls is typically modeled by a linear combination of the squared rotor speeds:
\begin{equation}\label{eq:thrust_mixing}
    \mb u = \mathcal{T} \bs \eta, \quad \mathcal{T}  = \Bigg [\begin{smallmatrix}
    c_T & c_T & c_T & c_T \\
    0 & c_T l & 0 & -c_T l\\
    -c_T l & 0 & c_T l & 0\\
    -c_Q & c_Q & -c_Q & c_Q 
    \end{smallmatrix} \Bigg], \quad \bs \eta = \Bigg[ \begin{smallmatrix} \eta_1^2 \\ \eta_2^2 \\ \eta_3^2 \\ \eta_4^2 \end{smallmatrix} \Bigg],
\end{equation}
where $c_T$ and $c_Q$ is the propeller thrust and torque coefficients, $l$ is the distance from the vehicle's center of gravity to the propeller axle,and $\eta_1, \dots \eta_4$ are the propeller rotation rates. This mapping enables the abstracted controls to be translated into propeller rotational rates.


\subsection{The learning process and data collection}
To collect the learning data set, we assume that $M_t$ data collection trajectories of length $T_t$, from initial conditions $\mb x_0^j \in \Omega, j=1,\dots,M_t$, can be executed under a nominal controller.   From each trajectory, $M_s = (T_t/\Delta t)$ state and control actions are sampled at a fixed time interval $\Delta t$, resulting in a data set
\begin{equation}
    \mathcal{D} = \bigg ( \big (\mb x_{j,k}, \mb x'_{j,k}, \mb u_{j,k} \big )_{k=0}^{M_s-1} \bigg )_{j=1}^{M_t},
\end{equation}
\noindent where $\mb x'$ is the state at the next timestep, i.e. $\mb x'_{j,k} = \mb x_{j,k+1}$.

Since the NMPC design requires continuous-time models to be discretized, we learn a discrete-time lifted bilinear model, thereby avoiding potential numerical differentiation and discretization errors. This is further motivated by the existence of discretization procedures that maintain stability properties and the bilinear structure of the original system, such as the trapezoidal rule with zero-order-hold \cite{Phan2012Discrete-timeModels, Surana2018KoopmanEstimation}. 

\begin{figure*}[!ht]
    \centering
        \subfloat[Autoencoder model \label{fig:autoencoder}]{%
      \includegraphics[height=3.75cm]{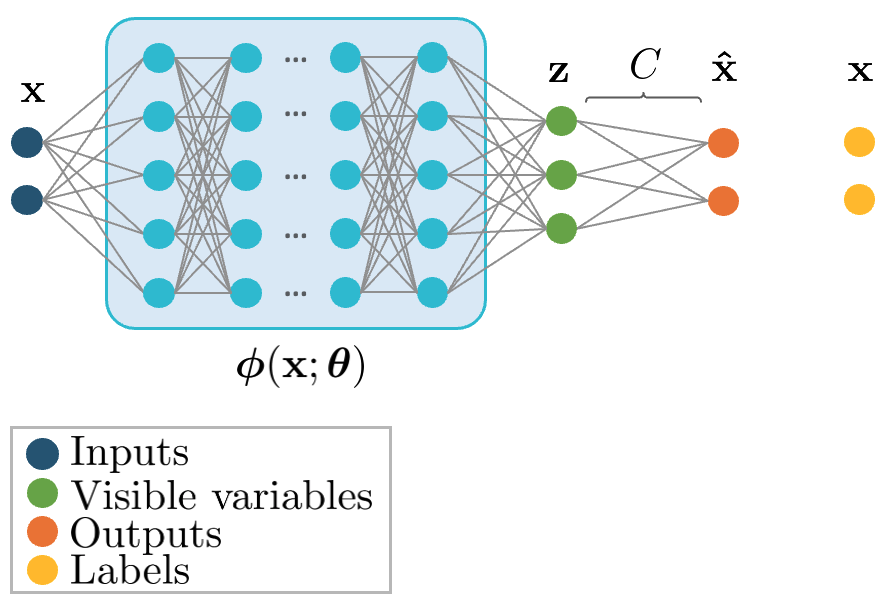}
    }
    \hfill
    \subfloat[State prediction model \label{fig:state_pred}]{%
      \includegraphics[height=3.75cm]{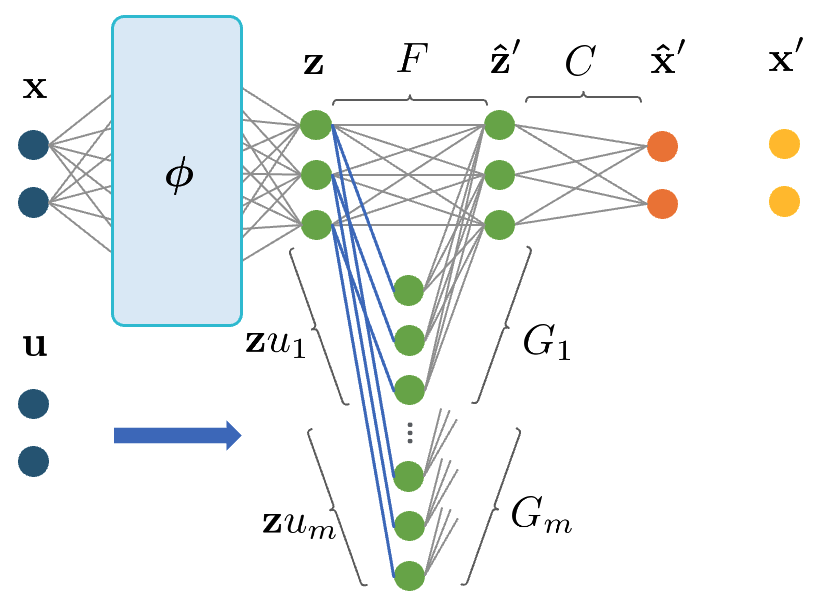}
    }
    \hfill
    \subfloat[Lifted state prediction model\label{fig:bilin_pred}]{%
      \includegraphics[height=3.75cm]{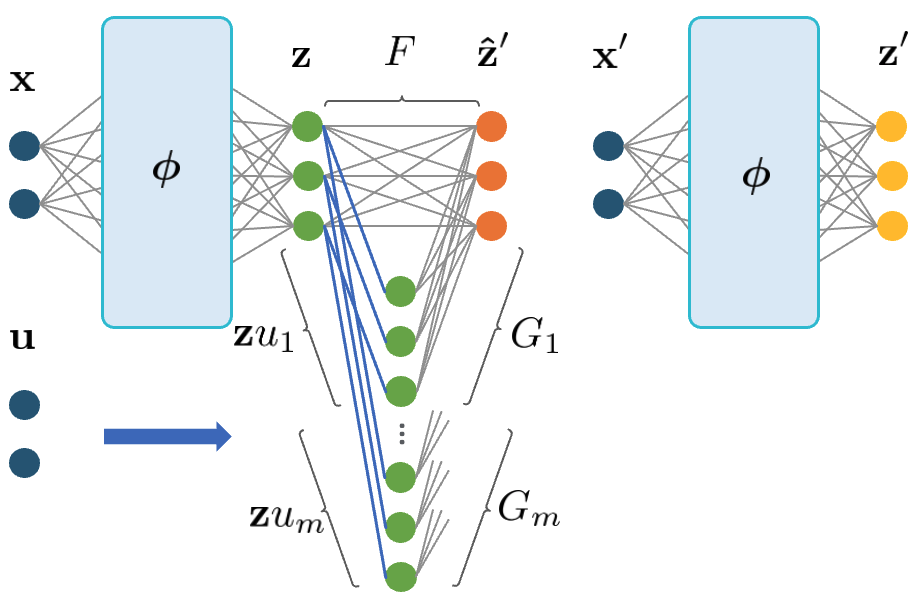}
    }
    \caption{Koopman neural network model architecture}
    \label{fig:dnn_architecture}
  \end{figure*}

\section{Joint learning of the Koopman dictionary and model}
\label{sec:learning}

To formulate the learning problem, we follow an analogous approach to the EDMD methods where we seek to use a dictionary of nonlinear transformations to lift the state to a higher dimensional space where the system's dynamics can be described by a bilinear model \cite{op_icra_21}. However, unlike classical EDMD, we parametrize the function dictionary by a neural network, and jointly learn the function dictionary and bilinear model matrices. As a result, we use $\theta$ to parametrize the neural network representing the function dictionary $\phi(\mb x;\bs \theta)$, and use $F, G_1, \dots, G_m, C^{\mb x}$ to represent the bilinear model matrices described in Section \ref{sec:prelims}.

The key idea of the learning method is that we can encode the function dictionary parametrization and the bilinear model structure in a single neural network. Then, once training is complete, the learned function dictionary and model matrices can be extracted to maintain the benefits of Koopman based learning methods outlined in Section \ref{sec:introduction}, while improving the prediction performance and/or reducing the dimension of the function dictionary. To achieve this, we formulate a loss function, $\mathcal{L}$ to be minimized by empirical risk minimization over all the input-output pairs in the training data set, $\mathcal D$:

\begin{align}\label{eq:dnn_loss}
\begin{split}
    &\mathcal{L}(\mb x, \mb u, \mb x') \!=\! \alpha \mathcal{L}_\text{rec}(\mb x, \mb x') \!+\! \mathcal{L}_\text{pred}(\mb x, \mb u, \mb x') \!+\! \beta \mathcal{L}_\text{kct}(\mb x, \mb u, \mb x'),\\ \\
    &\mathcal{L}_\text{rec}(\mb x, \mb x') \!=\! || \mb x \! - \! C^{\mb x} \bs \phi(\mb x; \bs \theta) ||^2\!,\\
    &\mathcal{L}_\text{pred}(\mb x, \mb u, \mb x') \!=\! || \mb x' \! - \! C^{\mb x}\bigg(\!\! F \bs \phi(\mb x; \bs\theta) - \!\!\!\!\!\! \sum_{i=1,\dots,m} \!\!\!\!\!\! G_i \bs \phi(\mb x;\bs \theta)\mb u_i \!\! \bigg) \! ||^2\!, \\
    &\mathcal{L}_\text{kct}(\mb x, \mb u, \mb x') \!=\! ||\bs \phi(\mb x';\bs \theta)\! - \!\!\bigg(\!\! F \bs \phi(\mb x; \bs \theta) - \!\!\!\!\!\!\sum_{i=1,\dots,m}\!\!\!\!\!\! G_i \bs \phi(\mb x;\bs \theta)\mb u_i \!\!\bigg) \!||^2\!,
\end{split}
\end{align}
where $\mathcal{L}_\text{rec}$ is the mean squared error (MSE) of the reconstruction loss when lifting the system's state using the encoder $\phi(\mb x;\bs \theta)$ and projecting back down to the original system state with the projection matrix $C^{\mb x}$, $\mathcal{L}_\text{pred}$ is the MSE of the one-step prediction error of the model when projected down in the original state space, and $\mathcal{L}_\text{kct}$ is the MSE of the one-step prediction error in the lifted space. Tunable hyperparameters $\alpha$ and $ \beta$ determine the weight of the prediction and KCT losses, respectively relative to the reconstruction loss. 

The learned model should realize good prediction performance in the original state space: minimizing $\mathcal{L}_\text{pred}$ encourages this goal. Loss terms $\mathcal{L}_\text{recon}$ and $\mathcal{L}_\text{kct}$ respectively promote accurate projection of the function dictionary to the original state space and an approximately bilinear dynamical system model in the lifted space. Loss $\mathcal{L}_\text{kct}$ promotes good prediction accuracy over multiple time steps, as multi-step prediction is performed in the lifted space before the result is projected to the original state space only at relevant times.

Figure \ref{fig:dnn_architecture} depicts the neural network architecture that implements loss (\ref{eq:dnn_loss}).
The autoencoder  (Fig. \ref{fig:autoencoder}) passes the system state $\mb x$ through the neural network, $\bs \phi(\mb x; \bs \theta)$, to obtain the lifted state $\mb z$. Subsequently, the lifted state is projected back to the original state space through the projection matrix $C^{\mb x}$, resulting in the reconstructed state, $\mb{\hat x}$, which is compared to the original state, $\mb x$, in $\mathcal{L}_\text{recon}$. The state prediction model (Fig. \ref{fig:state_pred}) passes  the system state through the neural network to obtain the lifted state, $\mb z$, before evolving the lifted state one time-step with the bilinear model based on $\mb z$ and $\mb u$ to get the one-step-ahead lifted state, $\mb z'$. The lifted state prediction is projected to the original state space to get the one-step-ahead state prediction, $\mb{\hat x}'$, which is compared to the true one-step-ahead state, $\mb x'$, in $\mathcal{L}_\text{pred}$. Finally, the lifted state prediction model (Fig. \ref{fig:bilin_pred}) follows the same forward pass through the same state prediction model to get the one-step-ahead lifted state prediction $\mb{\hat z}'$. This is compared to the true one-step-ahead lifted state $\mb z'$ in $\mathcal{L}_\text{kct}$, obtained by passing the true one-step-ahead state, $\mb x'$ through the neural network $\bs \phi$.

This approach can be implemented using modern neural network software packages with basis functions $\bs \phi$ modeled as a feedforward neural network having fully connected layers and $F, G_1, \dots, G_m, C^{\mb x}$ as single fully connected layers with no nonlinear activations. The entire model can be trained simultaneously using gradient-based learning algorithms applied to the loss function (\ref{eq:dnn_loss}). 

\section{Nonlinear model predictive control design}
\label{sec:nmpc}

\subsection{Design considerations}
We formulate the NMPC problem using the bilinear model identified by the Koopman framework \cite{op_icra_21}: 
\begin{align} \label{eq:koopman_nmpc}
    \begin{split}
        \min_{Z, U} \; & \sum_{i=0}^{N} \begin{bmatrix} C^{ \mb x} \mb z_{k} - \mb x^\text{ref}_k\\ \mb u_{k} \end{bmatrix}^T W_{k} \begin{bmatrix} C^{ \mb x} \mb z_{k} - \mb x^\text{ref}_k\\ \mb u_{k} \end{bmatrix}\\
        \text{s.t.} \quad &  \mb z_{k+1}\! =\! F \mb z_k\! +\! \sum_{j=1}^m G_i \mb z_k  \mb u_k^{(j)}, \; k\! =\! 0,\dots,N-1 \\
        & \mb c_l \! \leq \! C^{ \mb x}  \mb z_k \leq \mb c_u, \; \mb d_l \! \leq \! \mb u_k \leq \mb d_u, \;  k\! =\! 0,\dots,N,\\
        &  \mb \! z_0 = \! \bs \phi(\bs{\hat{\mb x}}) .
    \end{split}
\end{align}
where $\mb x^\text{ref}$ is the desired trajectory to be tracked.

Although (\ref{eq:koopman_nmpc}) models a quadratic cost function and linear state and actuation constraints, nonlinear objective and constraint terms can be included by adding them to the lifted state $ \mb z = [\bs \phi(\mb x; \bs \theta)^T, \bs \phi_\text{con}(\mb x)^T]^T  $. For example, if it is desired to enforce the constraint $\cos(x_1) \leq 0$, we can add $\phi_\text{con} = \cos(x_1)$ to the lifted state and enforce $ z_k^{(j)} \leq 0$ where $j$ is the index corresponding to $\phi_\text{con}$ \cite{Korda2018LinearControl}.

\subsection{Sequential quadratic programming}\label{sec:sqp}
The optimization constraint set is generally non-convex due to the bilinear term, $\mb z$, in the lifted dynamic model and control inputs, $ \mb u_1, \dots, \mb u_m$. As a result, we solve the optimization problem (\ref{eq:koopman_nmpc}) using sequential quadratic programming (SQP). The non-convex optimization problem (\ref{eq:koopman_nmpc}) is sequentially approximated by quadratic programs (QPs), whose solutions are Newton directions for performing steps toward the optimal solution of the nonlinear program (NLP). The sequence is initialized at a guessed solution, $(\mb z_0^{\text{init}}, \mb u_0^{\text{init}})$, at which the following QP is iteratively solved. The initial guess is updated at each iteration \textit{i} until converged:
\setlength{\belowdisplayskip}{2pt} 
\begin{align*} 
    \begin{split}
        &\min_{\Delta Z_i, \Delta U_i} \quad  \sum_{k=0}^N \begin{bmatrix} \Delta \mb z_{i,k}\\ \Delta \mb u_{i,k} \end{bmatrix}^T H_{i,k} \begin{bmatrix} \Delta \mb z_{i,k}\\ \Delta \mb u_{i,k} \end{bmatrix} + J_{i,k}^T \begin{bmatrix} \Delta \mb z_{i,k}\\ \Delta \mb u_{i,k} \end{bmatrix}\\
        \text{s.t.} \,& \Delta \mb z_{i,k+1} \!=\! A_{i,k} \Delta \mb z_{i,k} \!+\! B_{i,k}\Delta \mb u_{i,k} + \mb r_{i,k}, \, k\!=\!0,...,N\!\!-\!\!1,
            \end{split}
\end{align*}
\setlength{\abovedisplayskip}{0pt}
\begin{align} \label{eq:qp-mpc}
    \begin{split}
       & \mb c_l \leq C^{\mb x}(\mb z_{i,k}^\text{init} + \Delta \mb z_{i,k}) \leq \mb c_u, \, k\!=\!0,...,N, \quad \quad \, \\
        & \mb d_l \leq  \mb u_{i,k}^\text{init} + \Delta \mb u_{i,k} \leq \mb d_u, \, k\!=\!0,...,N,\\
        &  \Delta \mb z_{i,0} = \bs \phi(\mb{\hat{x}}; \bs \theta) - \mb z_{i,0}^\text{init} .
  \end{split}
\end{align}
    \setlength{\abovedisplayskip}{10pt} 
    \setlength{\belowdisplayskip}{10pt} 
    
where $\mb z_{i,k} = \mb z_{i,k}^\text{init} + \Delta \mb z_{i,k}$, $\mb u_{i,k} = \mb u_{i,k}^\text{init} + \Delta \mb u_{i,k}$, $H_{i,k}$ is the Hessian of the Lagrangian of the NLP (\ref{eq:koopman_nmpc}) and
\setlength{\belowdisplayskip}{0pt} 
\begin{align*}
    \begin{split}
        A_{i,k} &= F + \sum_{j=1}^m G_j (\mb u_{i,k}^{\text{init}})^{(j)}, \,\, B_{i,k} = \big [ G_1 \mb z_{i,k}^{\text{init}} \dots G_m \mb z_{i,k}^{\text{init}} \big ],
    \end{split}
\end{align*}
\setlength{\abovedisplayskip}{1.5 pt}
\begin{align} \label{eq:linearization}
    \begin{split}
        \mb r_{i,k} &= F\mb z_{i,k}^{\text{init}} + \sum_{j=1}^m G_j\mb z_{i,k}^{\text{init}} (\mb u_{i,k}^{\text{init}})^{(j)} - \mb z_{i,k+1}^{\text{init}},\quad \quad \quad \;\\
        J_{i,k} &= W_{i,k}\begin{bmatrix} \mb z_{i,k}^{\text{init}} - \mb z_{i,k}^{\text{ref}}\\{\mb u_{i,k}^\text{init}} - \mb u_{i,k}^{\text{ref}}\end{bmatrix}.
    \end{split}
\end{align}
    \setlength{\abovedisplayskip}{10pt} 
    \setlength{\belowdisplayskip}{10pt} 
    
As a result of the bilinear structure of the dynamics model, the linearization can be efficiently computed for a given initial guess by simple matrix multiplication and addition with the dynamics matrices of the Koopman model and the matrices containing the initial guesses of $Z_i^{\text{init}}, U_i^{\text{init}}$.



\begin{algorithm}[b]
\small
\SetAlgoLined
\textbf{Input:} reference trajectory $(X_i^{\text{ref}}\!\!\!, U_i^{\text{ref}})$, initial guess $(Z_i^{\text{init}}\!\!\!, U_i^{\text{init}})$\\
\While{Controller is running}{
Form $\mb r_{i,k}, A_{i,k}, B_{i,k}$ using (\ref{eq:linearization})\\
Get and lift current state, $\mb z_{i,0} = \bs \phi(\mb{\hat{x}}; \bs \theta)$\\
Solve (\ref{eq:koopman_nmpc}) to get the Newton direction $(\Delta Z_i, \Delta U_i)$\\
Update solution, 
$(Z_i, U_i)\! \leftarrow\! (Z_i^{\text{init}} \!+\! \Delta Z_i, U_i^{\text{init}}\!+\! \Delta U_i)$\\
Deploy first input $\mb u_0$ to the system\\
Construct $(Z_{i+1}^{\text{init}}, U_{i+1}^{\text{init}})$ using (\ref{eq:shift_procedure})
}
\caption{\cite{op_icra_21} Koopman NMPC (closed loop)}
\label{algo:koopman_mpc}
\end{algorithm}

\begin{figure*}[t]
    \centering
    \subfloat[Experiment set-up \label{fig:exp_set_up}]{%
      \includegraphics[height=5cm]{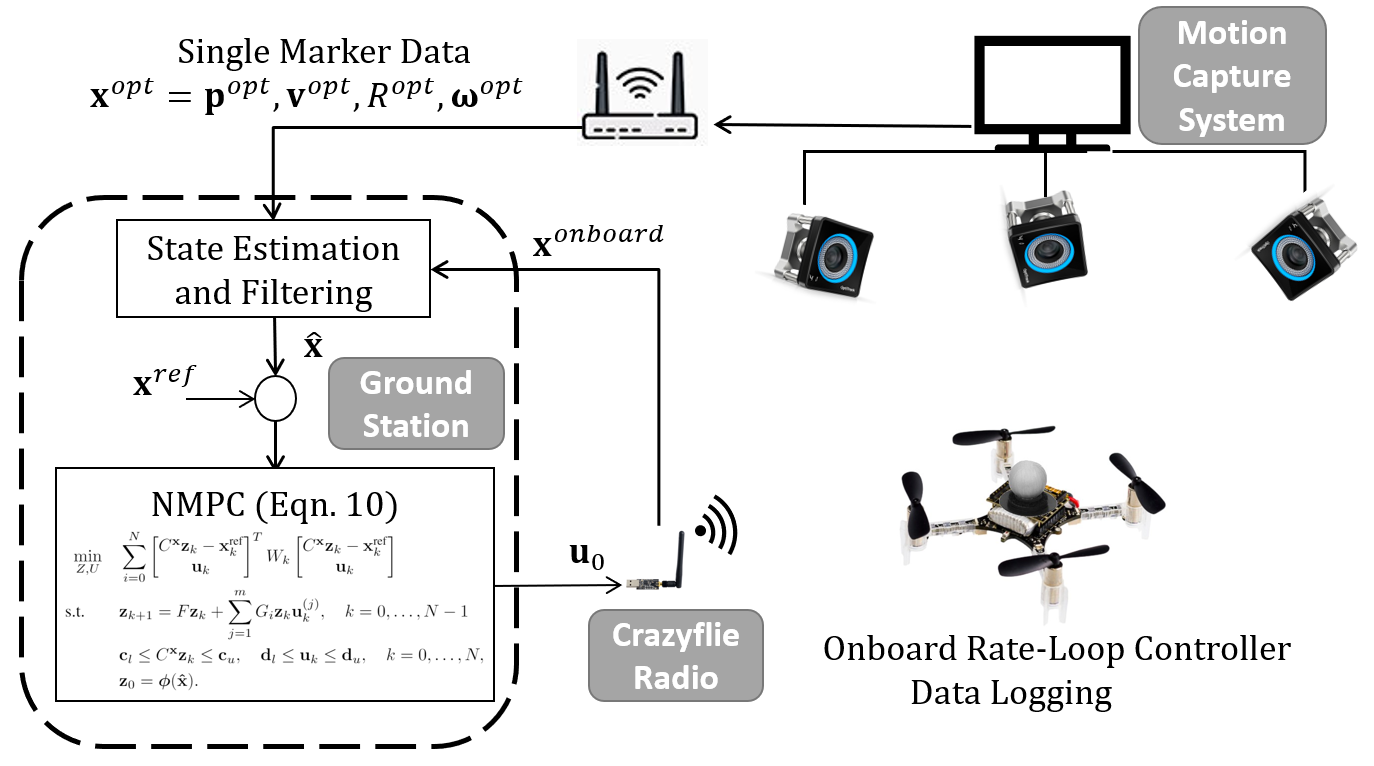}
    }
    \hfill
    \subfloat[``Figure 8" tracking with position depicted at selected times \label{fig:Figure_8_tracking_Experiment}]{%
      \includegraphics[height=5cm]{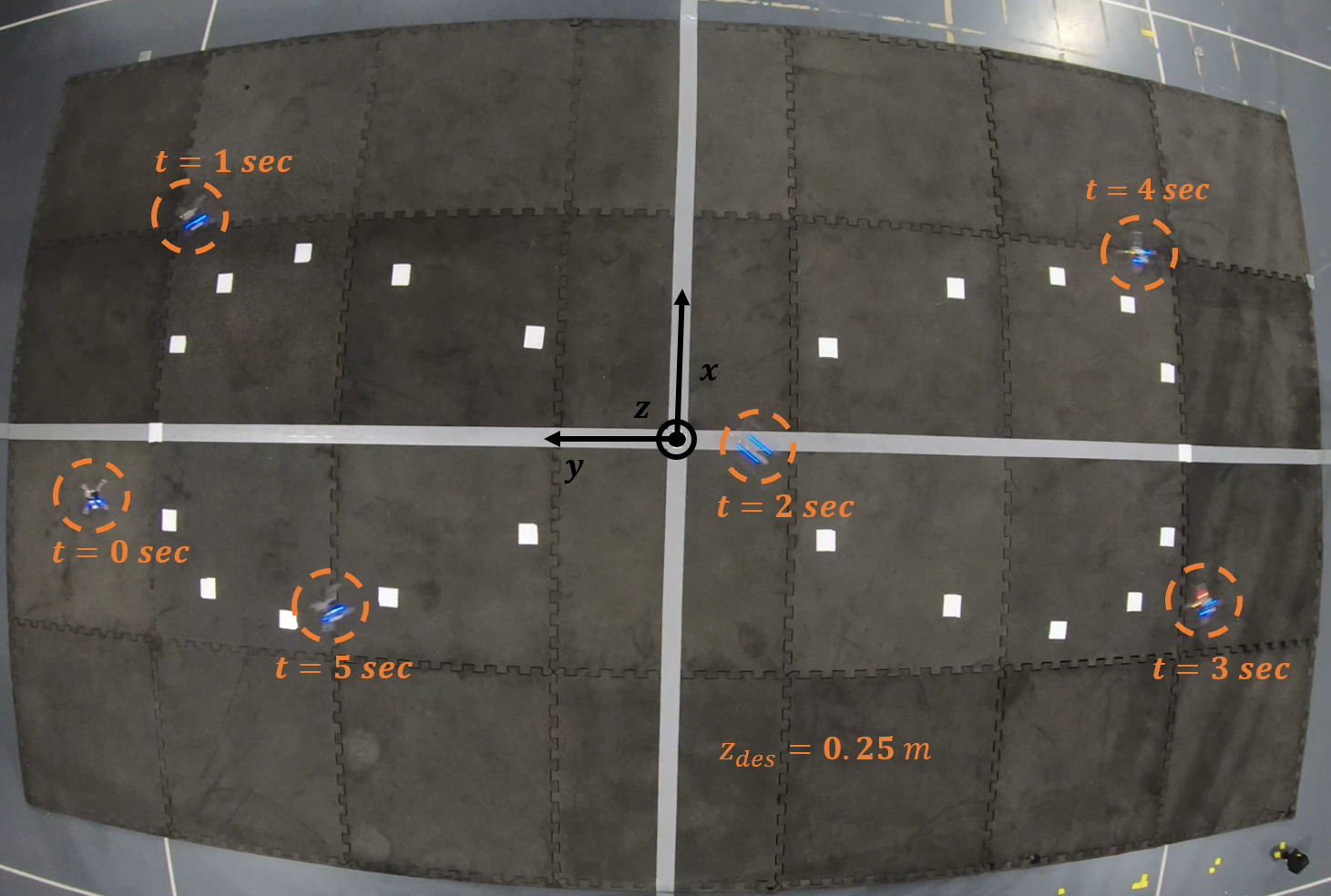}
    }
    \\
    \subfloat[Mean position coordinates (solid lines) +/- 3 std (shaded area) of 10 consecutive experiment runs with both controllers at 0.25 m altitude. \label{fig:exp_data}]{%
      \includegraphics[width=0.99\textwidth]{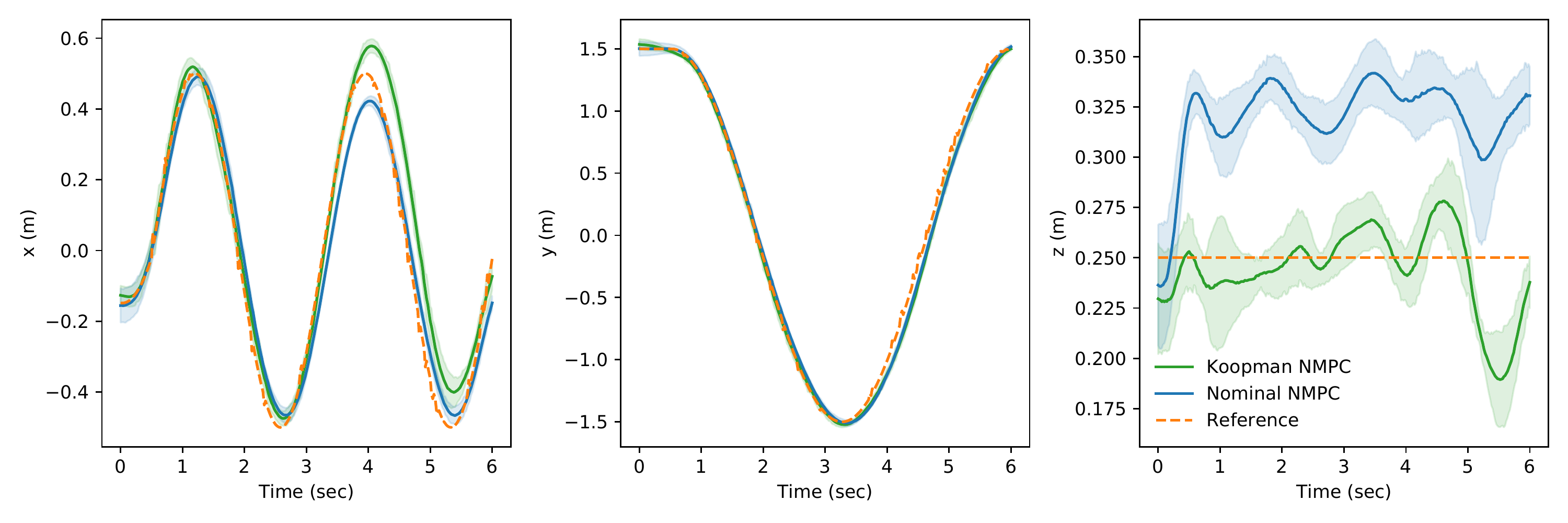}
    }
    \vspace{0.2cm}
    \caption{Hardware configuration and experimental results from ``Figure 8" trajectory tracking control experiment.}
\end{figure*}

\subsection{Warm-start of SQP at each timestep}

As discussed in Section \ref{sec:sqp}, the SQP algorithm requires an initial guess of the solution $Z_i^{\text{init}}, U_i^{\text{init}}$. Selecting an initial guess that is sufficiently close to the true optimal solution is essential for the algorithm to converge fast and reliably \cite{Gros2020FromIteration}. It is well known that the receding horizon nature of MPC can be exploited to obtain excellent initial guesses. At a time instant $i$, this can be achieved by \textit{shifting} the NMPC solution from the previous timestep $i-1$ and by updating the guess of the final control input. Under certain conditions, a locally stable controller enforcing state and actuation constraints can be designed allowing feasibility of the initial guess to be guaranteed \cite{Rawlings2012ModelDesign}. Typically, simpler approaches are taken, such as adding a copy of the final control signal and calculating the implied final state using the dynamics model
\begin{align} \label{eq:shift_procedure}
    \begin{split}
        &\mb u_{i,N-1}^{\text{init}}\! =\! \mb u_{i,N-2}^{\text{init}}, \; \mb u_{i,k}^{\text{init}}\! =\! \mb u_{i-1,k+1}, \; k \!=\! 0, \dots, N-2,\\
        &\mb z_{i,k}^{\text{init}} = \mb x_{i-1,k+1}, \quad k = 0, \dots, N-1,\\
        &\mb z_{i,N}^{\text{init}} = F\mb z_{i,N-1}^{\text{init}}+ \sum_{j=1}^m G_j \mb z_{i,N-1}^{\text{init}} \mb u_{i,N-1}^{\text{init},j}.
    \end{split}
\end{align}
If the previous solution $\mb z_{i-1}, \mb u_{i-1}$ is feasible, the shifted solution will also be feasible for all but the last timestep. 

\subsection{Koopman NMPC algorithm}
The initialization and closed loop operation of the controller can be summarized as follows (see Algorithm \ref{algo:koopman_mpc}). An initial guess of the state and control, $X^\text{init}_0, U^\text{init}_0$ is chosen and the state is lifted with the dictionary of functions at each timestep to obtain $Z^\text{init}_0 = \bs \phi(X^{init}_0; \bs \theta)$. Then, before task execution, the SQP algorithm with the Koopman QP subproblem (\ref{eq:koopman_nmpc}) is executed to convergence to obtain a good initial guess of the solution. Subsequently, in closed loop operation, the Koopman bilinear model is linearized along the initial guess. The current state is obtained from the system and lifted using the function dictionary, and then the QP subproblem is solved only once. Finally, the first control input of the optimal control sequence is deployed to the system, and the full solution is shifted one timestep and used as an initial guess at the next timestep.

\section{Experiments}
\label{sec:experiments}

We conduct experiments to demonstrate the performance of the proposed method on a quadrotor drone. To capture the nonlinearity in the quadrotor dynamics, we perform 6 second-long ``Figure 8" tracking maneuvers (see Fig. \ref{fig:Figure_8_tracking_Experiment}) at very low altitudes above the ground so that ``ground effect" corrupts the nominal model. The $x\!\!-\!\!y$ coordinates of the ``Figure 8" trajectory require high roll and pitch maneuvers and the low altitude flights highlight the effect of unmodeled dynamics, encapsulated in the $\mb f_\mb v$ and $\bs \tau_{\bs \omega}$ terms in \eqref{eq:dynamics}. 

\subsection{Implementation and experimental details}
The experiments are conducted with a commercially available \textit{Crazyfile 2.1} quadrotor. Global position is measured via an \textit{OptiTrack} motion capture system (tracking at 120 Hz) and fused with the onboard IMU data to get filtered state estimates. The all-up  quadrotor weight is 33.6 g, with a thrust to weight ratio of 1.7. Control commands are computed on a offboard computer and communicated to the drone over radio, see Fig. \ref{fig:exp_set_up}. All communication with the drone is done using the \textit{Crazyswarm} Python API \cite{Preiss2017Crazyswarm:Swarm}.

To collect training data, waypoints were sampled uniformly at random within $x,y,z \in [-0.3,0.3]\times[-0.3, 0.3]\times[0.1, 1.0]$ meters and tracked with a PID position controller for a total of 4 minutes. The estimated position, attitude, linear and angular velocites, motor pulse width modulation (PWM) signals, and battery voltage were collected at 500 Hz. The state estimates where downsampled and the PWMs and voltage averaged over each 0.02 s interval to obtain a dataset at 50 Hz, the target update rate for the controller. This smoothing process better captures motor PWM inputs, as the raw data showed significant variability within each 0.02 s period. Finally, we map PWMs and voltage to thrust using the model identified in \cite{Shi2020Neural-Swarm2:Interactions} (see Tab. 1, \cite{Shi2020Neural-Swarm2:Interactions}) and apply the thrust mixing  in (\ref{eq:thrust_mixing}) to get total thrust and torques around each axis, resulting in the inputs used in the learning process.

A discrete-time lifted-dimensional bilinear model is learned as described in Section \ref{sec:learning}. Thirty percent of the data set is held out for validation and testing and the hyperparameters are tuned to obtain good open loop prediction performance over 0.5 seconds, the same prediction horizon used in the model predictive controllers. The final parameters used are included in Tab. \ref{tab:training_hyperparams}. The neural network is implemented using \textit{PyTorch} \cite{Paszke2019PyTorch:Library}\footnote{\texttt{github.com/Cafolkes/koopman-learning-and-control}}. To simplify encoding the objective and constraints of the NMPC, the state itself is added to the lifted state, $\mb{z} = [\mb x^T, \bs \phi(\mb x; \bs \theta)^T]^T$, making the projection matrix known, $C^\mb x = [I, 0]$. This makes the reconstruction loss, $\mathcal{L_\text{recon}}$ in \eqref{eq:dnn_loss} redundant and we set $\alpha = 0$. 

Two controllers are implemented for the task. First, a nominal NMPC using the dynamic model (\ref{eq:dynamics}) with $\mb \tau_{\bs \omega}=\mb f_{\mb v}= \mb 0$ and the system parameters described in Tab. \ref{tab:controller_params}, is used as a performance baseline. Second, the Koopman NMPC described in Section \ref{sec:nmpc}, using the learned model, is implemented. Both controllers are coded in \textit{Python} using the \textit{OSQP} quadratic program solver \cite{Stellato2018OSQP:Programs}. The objective penalizes errors between the state and desired trajectory in $x, y, z$. The control inputs are constrained using the vehicle's mechanical limitations. The position error penalties, $Q = [q_x, q_y, q_z]$, and control effort penalties, $R = [r_T, r_{\tau_x}, r_{\tau_y}, r_{\tau_z}]$, and remaining control parameters are described in Tab. \ref{tab:controller_params}. The  control thrusts and desired attitude are calculated by the NMPCs and sent to the onboard drone controller, which decides the final motor control allocations. This architecture enables both controllers to cycle at 50 Hz while attitude tracking runs at a higher rate onboard the drone. 

\begin{figure}[t]
    \centering
    \includegraphics[width=\columnwidth]{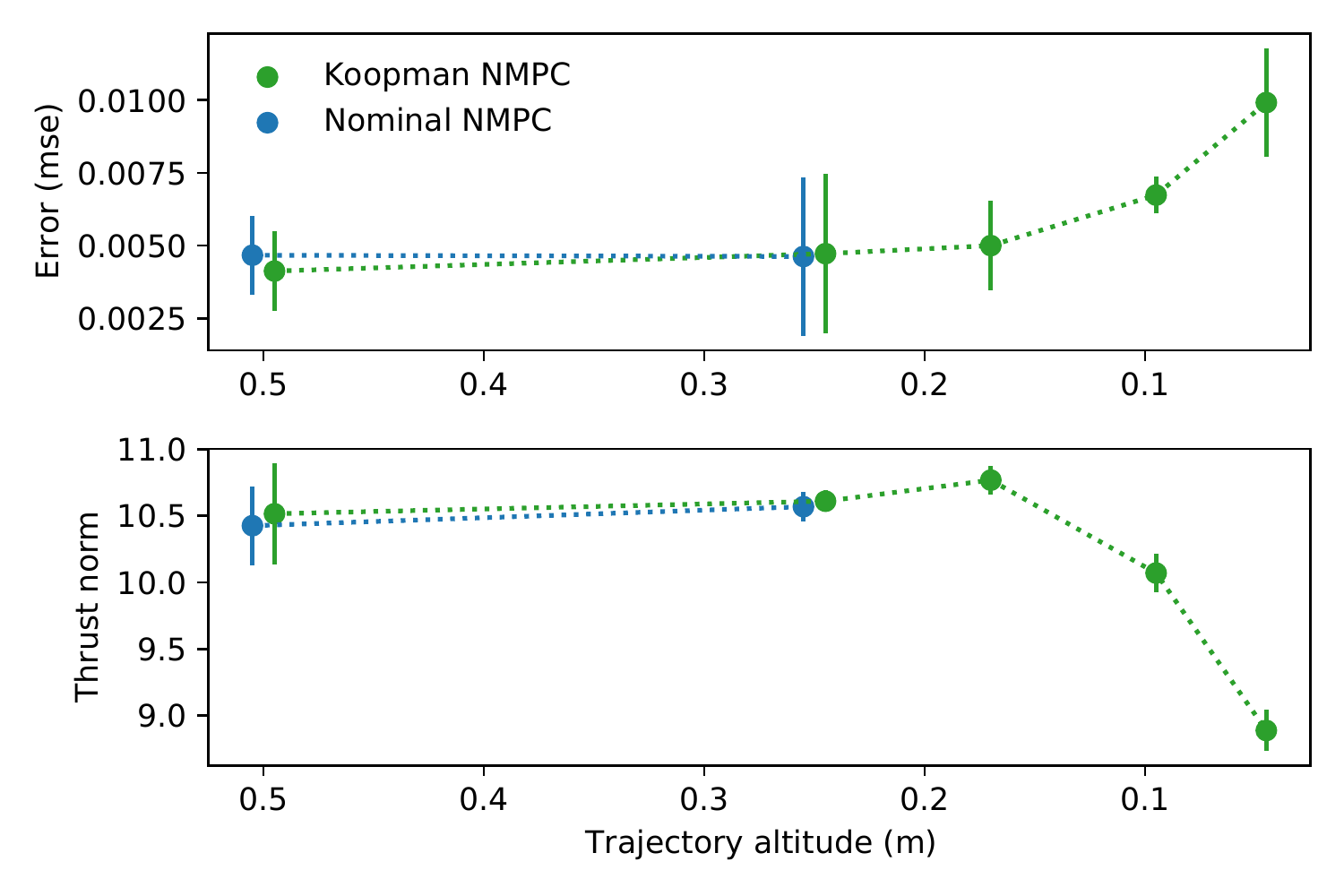}
    \caption{Average MSE and total thrust (dots) +/- 3 std (error bars) from 5 consecutive experiment runs with each of the controllers at decreasing altitudes. Stable flight not achieved by the nominal NMPC for altitudes below 0.25 m.}
    \label{fig:alt_comparison}
\end{figure}

\subsection{Results and discussion}

Fig. \ref{fig:exp_data} and Tab. \ref{tab:closed_loop} detail the tracking performance of the nominal and Koopman NMPC tracking a 6 second-long ``Figure 8" trajectory over 10 consecutive experiment runs. The yaw and pitch performance is comparable indicating accurate extraction of the roll-pitch-yaw dynamics. Furthermore, the large $z$-tracking error from the nominal NMPC induced by the ground effect is successfully captured by the Koopman learned dynamics. Note that the drone both starts and ends at zero velocity causing increased tracking error at the beginning and end of the trajectory.

To further study this conjecture, we executed 5  ``Figure 8" trajectories at decreasing altitudes, as shown in Fig. \ref{fig:alt_comparison}. At altitudes 0.25 m and higher, nominal and Koopman NMPCs tracking performance is similar. However, at lower altitudes nominal NMPC stability fails, leading to catastrophic crashes. In contrast, the Koopman NMPC completed all five tests, albeit with increased tracking error as the altitude setpoint approaches 0.05 m above the ground. We also note that the Koopman NMPC exploits the buoyancy in resulting from the ground effect at lower altitudes to significantly reduce the thrust needed to complete the task.

\begin{table}[t]
    \centering
    \vspace{0.23cm}
    \caption{Learning architecture and tuned hyperparameter values}
    \begin{tabular}{p{2.95cm}rp{2.95cm}r}
        \hline
         \# of encoder layers & 2 & Learning rate & 1e-3\\
         Layer width & 100 & KCT loss penalty, $\beta$ & 2e-1\\
         Activation function & tanh & $l_2$-regularization strength & 5e-4\\
         Total lifting dimension & 23 & \# of epochs & 200\\
         \hline
    \end{tabular}
    \label{tab:training_hyperparams}
    \vspace{0.14cm}
\end{table}

\begin{table}[t]
    \centering
    \caption{Crazyflie system properties and NMPC parameters}
    \begin{tabular}{p{1.35cm}rp{2.5cm}r}
        \hline
         Mass & 33.6 g & State error penalty, Q & [10, 10, 10]\\
         Max thrust & 57.0 g & Control penalty, R & [1, 1, 1, 1]\\
         $J_{xx}$ \cite{Landry2016AggressiveProgramming} & 1.7e-5 $\text{kg} \cdot \text m^2$ & NMPC prediction & \\
         $J_{yy}$ \cite{Landry2016AggressiveProgramming} & 1.7e-5 $\text{kg} \cdot \text m^2$ & horizon & 0.5 s\\
          $J_{zz}$ \cite{Landry2016AggressiveProgramming} & 2.9e-5 $\text{kg} \cdot \text m^2$ & Controller timestep & 0.02 s\\
         \hline
    \end{tabular}
    \label{tab:controller_params}
    \vspace{0.14cm}
\end{table}

\begin{table}[!t]
    \centering
    \caption{Tracking error and control effort of ``Figure 8" tracking experiment averaged over 10 experiment runs at 0.25 m altitude.}
    \begin{tabular}{p{5.2cm}|cc}
        \hline
         & Nominal & Koopman\\
         & NMPC & NMPC \\
         \hline
         Avg. position tracking error (mse) & 4.7e-3& 4.8e-3\\
         Avg. control effort (thrust norm) & 10.7 & 10.5\\
         \hline
    \end{tabular}
    \label{tab:closed_loop}
    \vspace{0.13cm}
\end{table}

\section{Conclusion}
\label{sec:conclusion}
The coupling of Koopman-based bilinear models and NMPC allows for real-time optimal control of robots that captures important nonlinearities, while allowing for critical state and control limits.
Function dictionary design is a key challenge in Koopman-based modeling and control methods. This paper presented a method to jointly learn a lifted Koopman bilinear model and KCT function dictionary strictly from data, enabling more compact models and/or better prediction performance compared to predefined function dictionaries. More compact models, i.e. lower lifting dimension, are crucial for robotic applications where real-time control is needed. Data-driven models are valuable in cases where first-principles modeling may be difficult. Our  quadrotor drone experiments demonstrate good prediction performance with a lifting dimension of only 23. The associated Koopman NMPC can match the performance of a NMPC based on the nominal model \eqref{eq:dynamics} far away from the ground and outperforms the nominal NMPC in near-ground regimes. Notably, the nominal controller is unable to maintain stable flight near the ground, whereas the Koopman NMPC maintains acceptable tracking performance down to 0.05 m altitude.

Future work includes deriving model error bounds for the learned model and exploiting the bounds to derive theoretical guarantees on controller performance. This can be achieved by extending our previous work to the bilinear model setting \cite{op_acc_21}.

\newpage
\bibliography{references} 
\bibliographystyle{ieeetr}
\end{document}